\documentclass[conference]{IEEEtran}
\IEEEoverridecommandlockouts
\usepackage{cite}
\usepackage{amsmath,amssymb,amsfonts}
\usepackage{mathtools}
\usepackage{graphicx}
\usepackage{xcolor}
\usepackage{bm}
\usepackage{bigints}
\usepackage{tikz}
\usepackage[export]{adjustbox}
\def\BibTeX{{\rm B\kern-.05em{\sc i\kern-.025em b}\kern-.08em
    T\kern-.1667em\lower.7ex\hbox{E}\kern-.125emX}}

\DeclarePairedDelimiter\abs{\lvert}{\rvert}%

\newcommand{\stencilIIIxIII}[9]
{ 
	\renewcommand{\arraystretch}{1.2}
	\begin{tabular}{|c|c|c|} \hline
		$\!#1\!$ & $\!#2\!$ & $\!#3\!$\\ \hline
		$\!#4\!$ & $\!#5\!$ & $\!#6\!$\\ \hline
		$\!#7\!$ & $\!#8\!$ & $\!#9\!$\\ \hline
	\end{tabular}
	\renewcommand{\arraystretch}{1.0}
}

\begin{document}
\title{CNN-based Euler's Elastica Inpainting with\\ 
Deep Energy and Deep Image Prior 
\thanks{
This work has received funding from the European Research Council (ERC)
under the European Union's Horizon 2020 research and innovation program
(grant agreement no. 741215, ERC Advanced Grant INCOVID).
}
}

\author{
\IEEEauthorblockN{Karl Schrader\IEEEauthorrefmark{1}, 
  Tobias Alt\IEEEauthorrefmark{1}, 
  Joachim Weickert\IEEEauthorrefmark{1}, 
  Michael Ertel\IEEEauthorrefmark{1}}
\IEEEauthorblockA{
  \IEEEauthorrefmark{1} Mathematical Image Analysis Group\\
  Faculty of Mathematics and Computer Science, Campus
  E1.7, Saarland University, 66041 Saarbr\"ucken, Germany\\
  Email: \{schrader, alt, weickert, ertel\}@mia.uni-saarland.de}
}

\maketitle

\begin{abstract}
Euler's elastica constitute an appealing variational image inpainting model. 
It minimises an energy that involves the total variation as well as the level 
line curvature. These components are transparent and make it attractive for 
shape completion tasks. However, its gradient flow is a singular, 
anisotropic, and nonlinear PDE of fourth order, which  is numerically 
challenging: It is difficult to find efficient algorithms
that offer sharp edges and good rotation invariance.
As a remedy, we design the first
neural algorithm that simulates inpainting with Euler's Elastica.
We use the deep energy concept which employs the variational energy as 
neural network loss. Furthermore, we pair it with a deep image prior 
where the network architecture itself acts as a prior.
This yields better inpaintings by steering the optimisation trajectory
closer to the desired solution.
Our results are qualitatively on par with state-of-the-art algorithms
on elastica-based shape completion. They combine good rotation 
invariance with sharp edges. Moreover, we benefit from the high efficiency 
and effortless parallelisation within a neural framework. Our neural 
elastica approach only requires $\bm{3 \times 3}$ central difference 
stencils. It is thus much simpler than other well-performing algorithms 
for elastica inpainting. Last but not least, it is unsupervised as it
requires no ground truth training data.
\end{abstract}

\begin{IEEEkeywords}
image inpainting, variational methods, CNNs, deep energy, deep image prior,
Euler's elastica
\end{IEEEkeywords}

\section{Introduction}
Inpainting \cite{MM98a,EL99a,BSCB00} aims at reconstructing 
missing regions of an image in a semantically plausible way.
In the last two decades, many inpainting techniques have been proposed
using model-based~\cite{GL14} ideas as well as deep  
learning~\cite{PKDD16,YLYS19}.
A particularly well-researched variational model goes back to 
Mumford~\cite{Mu94a} and utilises Euler's elastica~\cite{Eu44}. It 
minimises an energy with a total variation and a curvature term. Both 
terms offer a clear interpretation for shape completion tasks and lead 
to a particularly transparent model. 

\medskip
Unfortunately, the numerical minimisation
of this nonconvex energy is highly nontrivial, since its gradient 
flow is a singular, anisotropic nonlinear partial differential equation
(PDE) of order four. It is difficult to find efficient algorithms that 
produce sharp edges, offer a good approximation of rotation invariance, 
and can inpaint large gaps.  Numerous numerical ideas 
have been proposed to tackle this challenging task, including multigrid 
techniques~\cite{BC10}, discrete gradient methods~\cite{RLS18}, augmented 
Lagrangian approaches~\cite{THC11}, operator splitting~\cite{YK16}, and 
lifting concepts~\cite{CP19}; see \cite{KTZ19} for an overview and
additional references. To the best of our knowledge, however, no deep 
neural networks have been used for solving elastica inpainting. In view 
of the success of convolutional neural networks (CNNs) \cite{GBC16} in 
all visual computing fields as well as in numerical analysis, this is 
surprising. The goal of our paper is to close this gap.
 
\medskip
\noindent
\textbf{Our Contributions.}
Our approach focuses on \textit{deep energies}~\cite{GFE21}. It uses 
the variational energy as loss function for training a neural network
without ground truth data. For digital images, this only requires to 
discretise the energy. The task of its numerical minimisation 
via gradient descent is delegated to  the neural network. 
In this way we benefit from its powerful optimisation 
algorithms and an effortless parallelisation on GPUs. It should be noted
that the discretisation of the elastica energy only involves first and 
second order derivatives. This is far more pleasant than discretising a 
fourth order gradient flow. We show that $3 \times 3$ central differences  
with optimised rotation invariance are sufficient for this task.

\medskip
While our discrete energy is simple, it cannot penalise checkerboard-like 
artefacts. As a remedy, we employ a second neural concept, namely 
\textit{deep image priors}~\cite{UVL18}: This approach reparametrises an 
image as the output of a neural network and has a regularising effect. 
It efficiently prevents the introduction of unnatural artefacts into the 
resulting image. 

\medskip
Our experiments show that both neural components are essential for 
obtaining a good minimiser of the elastica energy. We even reach the 
quality of a sophisticated state-of-the-art algorithm. Our approach is  
efficient, offers sharp results with good rotation invariance, and can 
bridge large gaps.

\medskip
\noindent
\textbf{Related Work.}
Combinations of Euler's elastica and neural concepts are rare and
restricted to areas beyond image inpainting.
The use of elastica for supervised classification is investigated 
in~\cite{LXWH15}. There, they act as a regulariser on the level lines of a 
learned classifier. An elastica-based segmentation model where the 
minimiser is predicted by a neural network was explored in~\cite{CLZZ20}.

\medskip
Inpainting models that rely on deep learning can produce visually realistic 
images; see e.g.~\cite{PKDD16,YLYS19}. 
However, they require tremendous amounts of training data, and uncovering the 
learned model is infeasible due to the large number of trainable parameters. 
The concept of deep energies~\cite{GFE21} follows a different philosophy by 
proposing to use classical variational energy models as loss functions for 
neural networks. Deep energy models require no ground truth training data, 
and the mathematical model is fully defined by the energy. 

\medskip
Deep image priors~\cite{UVL18} have shown how the architecture of a neural 
network can act as a regularising prior on its output~\cite{DKMO20}. They 
preserve natural visual objects while attenuating noise. Therefore, 
their optimisation trajectories pass close by the desired solution in 
tasks such as noise or artefact removal.
Many works build upon this concept and propose different strategies for 
stopping the iteration as close as possible to the desired solution; 
see e.g.~\cite{WLZC21} and the references therein.

\medskip
\noindent
\textbf{Organisation of the Paper.}
In Section~\ref{sec:review}, we briefly review Euler's elastica
model for inpainting. Afterwards, in Section~\ref{sec:ours}, we introduce our
neural algorithm. We evaluate it in Section~\ref{sec:experiments} and
present our conclusions in Section~\ref{sec:conclusions}.

\section{Review: Elastica Inpainting}\label{sec:review}
Let $f: \Omega \to \mathbb{R}$ denote a continuous greyscale image that is 
only known on an \textit{inpainting mask} $K$, a subset of the rectangular 
image domain~$\Omega$. Inpainting aims at reconstructing $f$ in the 
\textit{inpainting domain} $\Omega\setminus K$. Euler's elastica obtain
such a reconstruction $u$ as a minimiser of the energy  
\begin{equation}\label{eq:elastica_cont}
E(u) \;=\; \int\limits_{\Omega\setminus K} \abs{\bm \nabla u}
\,\Big(b + \left(1-b\right) \kappa^2\Big)\,dx\,dy.
\end{equation}
On the inpainting mask $K$, we enforce $u=f$. 
The energy~\eqref{eq:elastica_cont} combines the total variation 
(TV)~\cite{ROF92} penaliser $\abs{\bm \nabla u}$ with the level line 
curvature $\kappa = 
\bm{\nabla}^\top\!\left(\frac{\bm\nabla u}{\abs{\bm \nabla u}}\right)$,
where $\bm{\nabla}=(\partial_x,\partial_y)^\top$ is the nabla operator, and
$\abs{\,\cdot\,}$ denotes the Euclidean norm.
The balance between the two components is steered by a parameter 
$b\in [0, 1]$. It can produce results from pure contour length minimisation
(for $b=1$) to pure curvature minimisation ($b=0$). Both components are 
highly desirable and psychophysically relevant for shape 
completion~\cite{Ka79}.

\medskip
Although the elastica energy \eqref{eq:elastica_cont} is transparent, it
involves many inherent problems: $\abs{\bm \nabla u}$ is nondifferentiable 
in $\bm{0}$,
and dividing by it within the curvature term may create singularities.
Moreover, since $\abs{\bm \nabla u}\,\kappa$ is the second derivative
of $u$ in level line direction, the elastica energy is highly
anisotropic. It is also nonconvex and may thus have 
numerous local minimisers. Any minimiser is a steady state of the 
gradient flow PDE of the energy. Since the energy~(\ref{eq:elastica_cont}) 
is nonquadratic, its gradient flow is nonlinear. Moreover, 
(\ref{eq:elastica_cont}) involves derivatives up to order two.
This creates a gradient flow of order four; see~\cite{SKC02} 
for an explicit formula. Fourth order PDEs are numerically much harder 
to solve than the widespread second order ones. Other challenges 
such as nondifferentiability, singular behaviour, and anisotropy are 
inherited. 

\medskip
As already mentioned, these problems directly lead 
to numerous numerical challenges. They have inspired many researchers
to develop highly sophisticated algorithms for elastica inpainting. 
We show that the concepts of deep energies and deep priors help us
to make these problems more manageable and lead to a simple and 
well-performing algorithm.

\section{Solving Elastica with Deep Learning}\label{sec:ours}
Our paper follows the \textit{discretise then optimise} paradigm.
Discretising the energy rather than its fourth order gradient flow 
allows us to restrict ourselves to derivatives up to order two.
Moreoever, we know that the resulting algorithm minimises a discrete
energy, which we can use to monitor its success. Most importantly,
however, we can exploit the capabilities of neural networks to 
minimise difficult nonconvex energies in an efficient way.

\subsection{Discretising the Elastica Energy}

The components of the energy~\eqref{eq:elastica_cont} can be spelled out as  
\begin{align}
 \abs{\bm \nabla u} &\;\approx\; \sqrt{u_x^2 + u_y^2 + \varepsilon^2}\,, \\
 \kappa &\;\approx\; \frac{u_y^2u_{xx} -2u_xu_yu_{xy}+ u_x^2u_{yy}}
        {(u_x^2+u_y^2+\varepsilon^2)^\frac{3}{2}}\,,
\end{align}
where subscripts denote partial derivatives. 
Regularising $\abs{\bm \nabla u}$ with $\varepsilon>0$ avoids
nondifferentiability and division by zero.

\medskip
We obtain discrete images $\bm u, \bm f$ by sampling the continuous functions 
$u, f$ on a uniform grid with distance $h$. We represent images as vectors 
by stacking the grey values into a column vector. Moreover, we consider a 
discrete binary inpainting mask $\bm c$ as an indicator image of the set 
$K$. A value of $1$ marks known data, and $0$ indicates the inpainting domain.

\medskip
Finite difference discretisations of derivatives are always a compromise
between many criteria: Ideally they are 
simple, offer good rotation invariance, do not introduce artificial blur,
and have a high approximation quality for all frequencies. 
In practice, improving the performance w.r.t.~one criterion often comes 
at the expense of another.
Our discretisation prioritises \textit{simplicity} and good 
\textit{rotation invariance,} since both are frequent weaknesses of 
existing elastica algorithms. 
We shall also see that it does not create blurry edges. 
Its main drawback are checkerboard artefacts in the highest grid frequency. 
We will cure them with a deep image prior. 

\medskip
For \textit{simplicity}, we 
approximate all derivatives by finite differences on a $3 \times 3$ 
stencil, which allows consistency
order two. To guarantee that they all fit together, 
we obtain them jointly from the coefficients of a weighted least squares 
regression polynomial of order two. We achieve good \textit{rotation 
invariance} with the tensor product of the binomial weights 
$[\frac{1}{4},\frac{1}{2},\frac{1}{4}]$ in $x$- and $y$-direction,
which approximates a rotationally invariant 2-D Gaussian. 
This yields
\begin{equation} \label{eq:stencil1}
\partial_x \approx \frac{1}{8h}\,
   \stencilIIIxIII{-1}{0}{1}{-2}{0}{2}{-1}{0}{1} ~, \quad
\partial_{xx} \approx \frac{1}{4h^2}\,
   \stencilIIIxIII{1}{-2}{1}{2}{-4}{2}{1}{-2}{1}
\end{equation}
and corresponding stencils in $y$-direction. Note that our approximations
for $\partial_x$ and $\partial_y$ coincide with Sobel operators, which are 
well-known for their rotation invariance. For the mixed derivative
we obtain (with $y$-axis oriented upwards)
\begin{equation} \label{eq:stencil2}
\partial_{xy}\approx \frac{1}{4h^2}\,
   \stencilIIIxIII{-1}{0}{1}{0}{0}{0}{1}{0}{-1} ~.
\end{equation}
With these stencils and by replacing the integral by a summation
over all pixels $i$ in the inpainting domain (i.e.~in locations~$i$ 
with $c_i=0$), we arrive at our discrete counterpart 
of~\eqref{eq:elastica_cont}:
\begin{equation}\label{eq:elastica_loss}
  E\!\left(\bm u\right) \;=\;
   \sum_{\substack{i}}\,(1-c_i)\:\abs{\bm \nabla u}_{i}\,
   \Big( b + \left(1-b\right) \kappa_{i}^2 \Big)\,.
\end{equation}

\subsection{Minimisation via Deep Energies}
To minimise this energy with advanced variants of gradient
descent, we use it as a loss function in a neural network. Golts et 
al.~\cite{GFE21} call this a \textit{deep energy}. Modern deep 
learning frameworks allow us to implement the discrete energy $E(\bm u)$ 
and use backpropagation \cite{RHW86} to evaluate $\bm\nabla_{\bm u}E(\bm u)$.
With this we can perform gradient descent:
\begin{equation}
\bm u^{k+1} = \bm u^k -\tau\bm\nabla_{\!\bm u^k}E(\bm u^k),
\label{eq:gradient_descent}
\end{equation}
where $\tau$ denotes the step size and superscripts the iteration level.
For Euler's elastica, this frees us from the need to find a good 
discretisation of a fourth-order PDE. Instead, only 
the energy which contains derivatives up to second order  
must be discretised. Its gradient is efficiently computed by the network 
by exploiting the parallelism of GPUs.
Note that so far, our network is used as a pure optimisation tool.
It has no trainable weights apart from the iteratively inpainted 
image~$\bm u^{k}$.

\subsection{Regularisation with a Deep Image Prior}

By design, our finite difference stencils 
\eqref{eq:stencil1}--\eqref{eq:stencil2} are simple and offer good 
rotation invariance. However, it is easy to see that the discrete
elastica energy does not penalise oscillations with the highest grid
frequency, where both Sobel operators return zero. As a result, 
checkerboard artefacts can appear. We 
avoid them with the regularising properties of a deep image 
prior~\cite{UVL18,DKMO20}: It has been shown that typical neural
networks converge rapidly towards natural images while attempting 
to avoid unnatural artefacts. As a consequence, our resulting model 
first recovers an image that fulfils the desired elastica 
properties, and may introduce undesirable artefacts afterwards. 
Thus, stopping the minimisation at the right time is crucial.

\medskip
We impose a deep prior on $\bm u$ by replacing it by the 
output of a network $\bm u = \mathcal{N}(\bm c, \bm f, \bm \theta)$ 
with mask $\bm c$ and known grey values~$\bm f$ as inputs, and parametrised by 
weights $\bm\theta$. 
Instead of searching for the minimiser $\bm u$ directly, we are 
now solving for the weights $\bm \theta$ which minimise 
$E(\mathcal{N}(\bm c, \bm f, \bm \theta))$.
Gradient descent gives 
\begin{equation}
 \bm \theta^{k+1} = \bm \theta^k -\tau\bm\nabla_{\!\bm{\theta}^k}
 E(\mathcal{N}(\bm c, \bm f, \bm \theta^k)) .
\end{equation}
Note the close connection to the explicit 
scheme~\eqref{eq:gradient_descent}, 
as the gradient w.r.t. the weights is computed via the chain rule. With
$\bm u^k \coloneqq \mathcal{N}(\bm c, \bm f, \bm \theta^k)$ and 
the Jacobian $\bm \nabla_{\!\bm \theta^k} \bm u^k$, this reads
\begin{equation}
\bm \nabla_{\!\bm \theta^k} E(\bm u^k) 
=  (\bm \nabla_{\!\bm \theta^k}\bm u^k) \,
   \bm \nabla_{\!\bm u^k} E(\bm u^k). \label{eq:chain_rule}
\end{equation}
The gradient $\bm\nabla_{\!\bm u^k}E(\bm u^k)$ is calculated first before 
being distributed onto the contributing weights $\bm \theta^k$.

\medskip
Our experiments will demonstrate that this deep prior regularisation
efficiently attenuates checkerboard artefacts. Moreover,
by avoiding irrelevant local minima, it brings us
closer to the desired solution after a reasonable number
of iterations.

\subsection{Our Inpainting Network Architecture}\label{sec:network} 
\begin{figure*}
\centering
\includegraphics[]{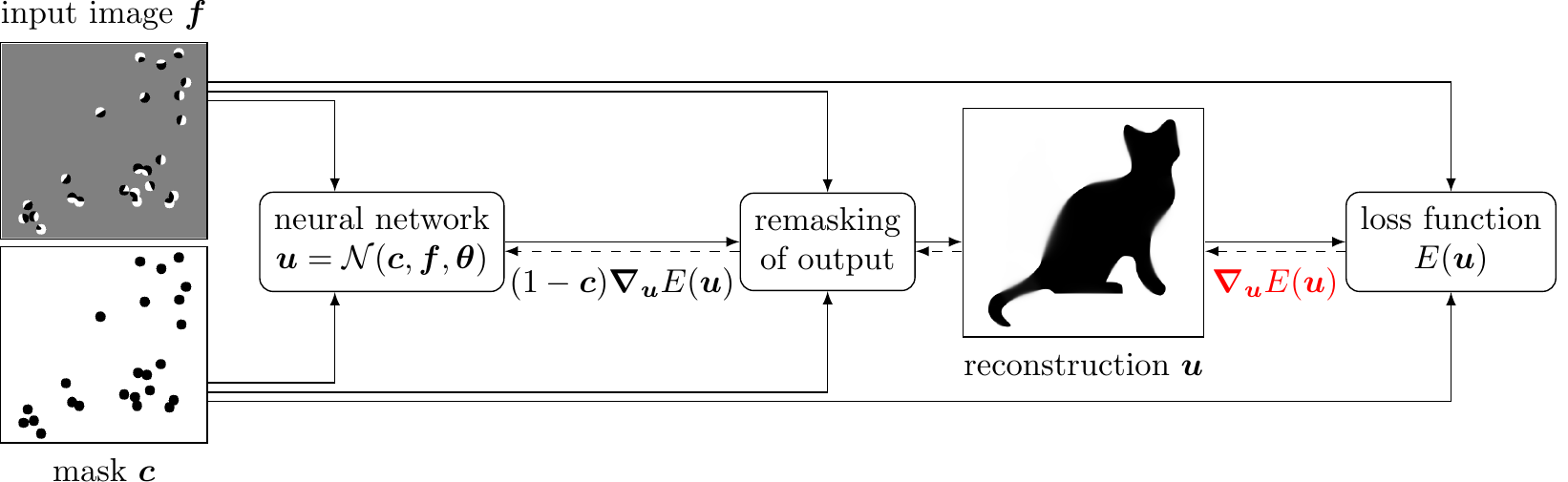}
\caption{The full network architecture. The flow of gradients during 
	backpropagation is depicted with dashed lines.
	Notice that $\bm \nabla_{\bm u} E(\bm 
	u)$ is still computed as part of the backpropagation, which shows 
	the close connection to gradient descent.}
\label{fig:full_flow}
\end{figure*}

In Figure \ref{fig:full_flow} we outline the full pipeline of our neural 
algorithm for elastica inpainting. 
As the energy is only considered within the inpainting domain, we 
remask the network output $\bm u$ with the known data $\bm f$. 
The final reconstruction is passed to the deep energy loss function 
which evaluates its quality.

\medskip
Our architecture consists of a small gated U-net~\cite{RFB15, YLYS19}.
Ulyanov et al.~\cite{UVL18} found that variants of U-nets perform well as 
deep priors for tasks like inpainting or artefact removal.
Furthermore, gated U-nets were used successfully for free-form 
inpainting as in our setting~\cite{YLYS19}, and we confirm their suitability 
as deep priors in our experiments.

\medskip
The defining feature of U-nets is their shape: The left part consists of
a sequence of convolutions and downsampling operations, resulting in features on
different scales. In a similar way, the right side repeatedly convolves 
and upsamples the features, and it concatenates them with features of the 
same scale from the downsampling pass. The gated U-net enhances the 
architecture by jointly evolving features and masks in each gated 
convolutional layer; see \cite{YLYS19} for a detailed explanation.

\section{Experiments}\label{sec:experiments}

In this section we first present an ablation study which demonstrates that
both neural components of our approach are essential. Afterwards we
evaluate its performance for inpainting of natural images and shape
completion tasks.

\subsection{Experimental Setup}
We prefer standard U-nets for fast inpaintings of natural images, and 
gated ones for shape completion with highest quality. 
Depending on the size of the image, three or four scales with two 
convolutional layers each are used. 
We start with up to 28 channels at the finest resolution and double them with 
each downsampling. Following~\cite{YLYS19}, dilated convolutions on the 
coarsest scale of the gated U-nets are included. The inpainting region of 
the masked image is initialised with random uniform noise in the same 
range $[0,1]$ as the image. 

\medskip
The network is trained with the Adam optimiser~\cite{KB15}. 
We start with a learning rate of $\tau = 0.001$ for natural images and 
$\tau=0.00005$ for shape completion and decrease it in later epochs. 
The parameter $b$ is chosen as to minimise the mean absolute error~(MAE) 
w.r.t.~the ground truth. We use the MAE since it better reflects our visual 
impression than the mean squared error (MSE), while remaining mathematically 
more transparent than pure perceptual measures.

\subsection{Ablation Study}

\begin{figure}
\centering
\footnotesize
\setlength{\tabcolsep}{2.8pt}
\begin{tabular}{cc}
          & deep energy only,   \\
 original &  without deep prior \\[1mm]
\includegraphics[width=27mm, frame]
   {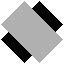} &   
\includegraphics[width=27mm, frame]
   {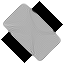} \\[2.8pt]
\includegraphics[width=27mm, frame]
   {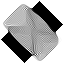} &
\includegraphics[width=27mm, frame]
   {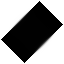} \\[1mm]
 ditto, with additional & deep energy with\\
 10,000,000 iterations  & gated U-net deep prior 
\end{tabular}
\caption{Comparison of results with and without deep prior ($b=0.001, 
\varepsilon=0.0001$, and $60,000$ iterations). For simplicity, the 
inpainting region is initialised with the average gray value.
The learning rate starts at $\tau = 0.00004$ and is 
halved every $20,000$ iterations. The optimisation producing the 
third image ran for an additional $10,000,000$ iterations at the final 
learning rate.}  
\label{fig:vs_gd}
\end{figure}

The ablation study in Figure~\ref{fig:vs_gd} shows that the deep energy
alone is insufficient for obtaining the desired inpainting result,
regardless of the number of iterations. Incorporating the deep image prior
is essential. It helps to escape from bad local minima: From the third to 
the fourth image, the discrete energy drops from $2.22$ to the remarkably 
low value of $0.08$.

\subsection{Inpainting of Natural Images}

\begin{figure*}
\centering
\footnotesize
\setlength{\tabcolsep}{2.8pt}
\begin{tabular}{ccccc}
original \textit{trui} image & 
random mask & 
1,000 iterations & 
{\color{red} 6,000 iterations} & 
200,000 iterations\\
$256 \times 256$ & 
$10$ \% density & 
MAE: $2.4\cdot10^{-2}$ & 
{\color{red} MAE: $1.6\cdot10^{-2}$} & 
MAE: $3.1\cdot10^{-2}$\\[1mm]
\includegraphics[width=33mm, frame]
  {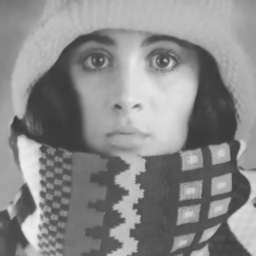} &   
\includegraphics[width=33mm, frame]
  {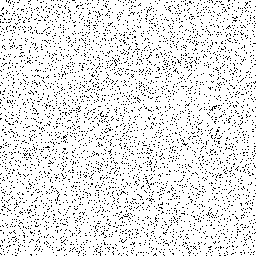} &
\includegraphics[width=33mm, frame]
  {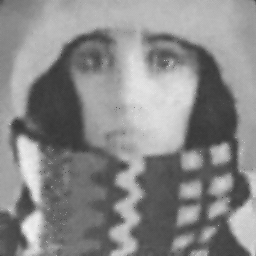} &
\includegraphics[width=33mm, frame]
  {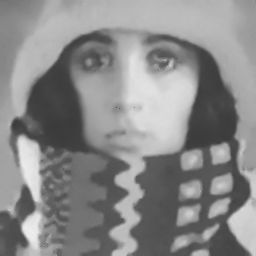} &		
\includegraphics[width=33mm, frame]
  {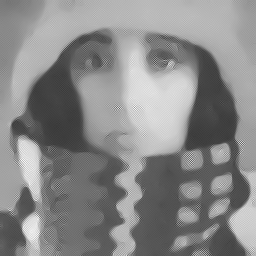} \\[2.8pt]
\includegraphics[width=33mm, frame]
  {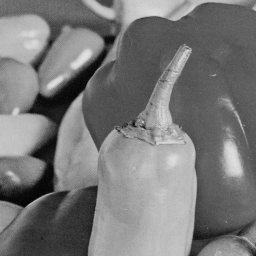} &   
\includegraphics[width=33mm, frame]
  {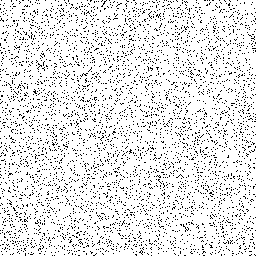} &
\includegraphics[width=33mm, frame]
  {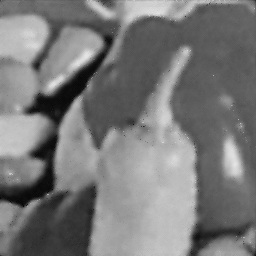} &
\includegraphics[width=33mm, frame]
  {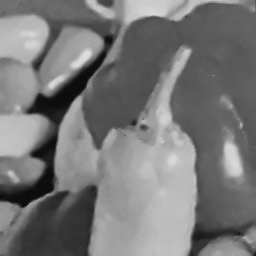} &		
\includegraphics[width=33mm, frame]
  {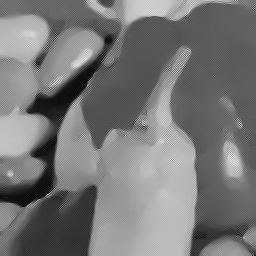}\\[1mm]
original \textit{peppers} image & random mask & 1,000 iterations & 
{\color{red} 7,000 iterations} & 
200,000 iterations\\ 
$256 \times 256$ & $10$ \% density & 
MAE: $2.5\cdot10^{-2}$ & 
{\color{red} MAE: $1.9\cdot10^{-2}$} &
MAE: $3.1\cdot10^{-2}$
\end{tabular}
\caption{Inpainting results for \textit{trui} and \textit{peppers} 
  ($b=0.175$, $\,\varepsilon=0.005$, $\,\tau = 0.001$). 
  After 6000--7000 iterations, a good reconstruction with low
  mean average error (MAE) is obtained. Additional iterations produce 
  checkerboard artefacts (digital zoom recommended).} 
\label{fig:progression}
\end{figure*}
\begin{figure}
\centering
\includegraphics[width=0.88\columnwidth]
  {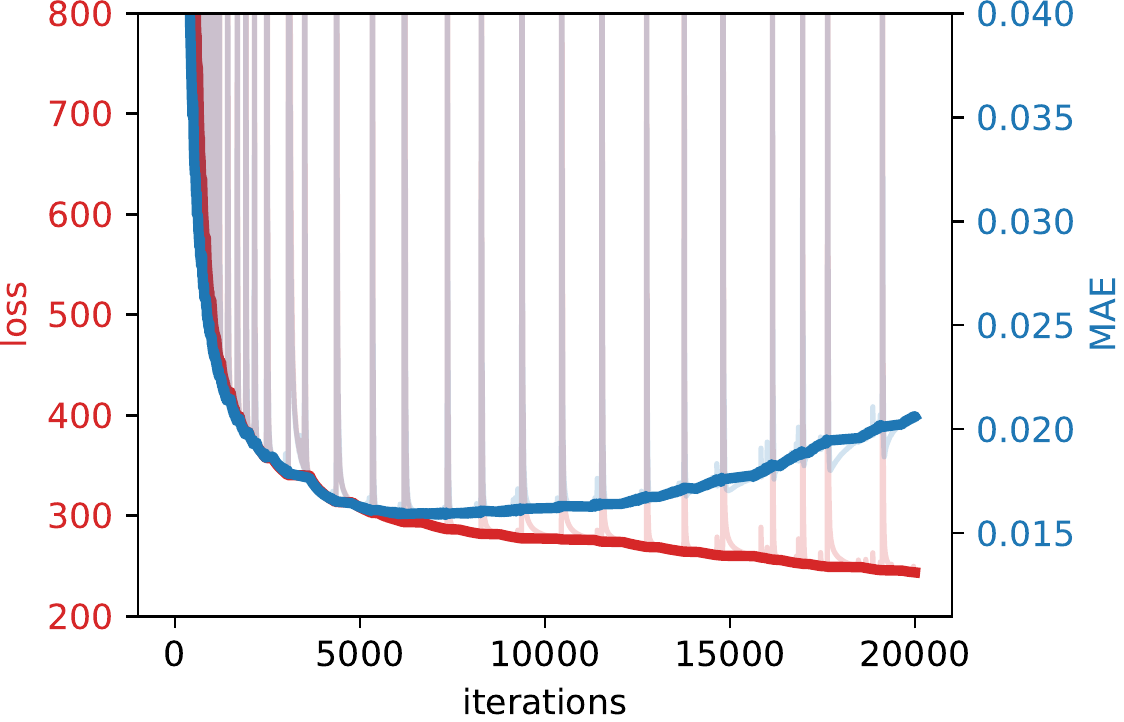}

\vspace{-2mm}
\caption{Energy and reconstruction error over time corresponding to   
	\textit{trui} in Figure~\ref{fig:progression}.
  Spikes were filtered from the bold lines to make the general trends 
  more recognisable. They are caused by brief, very large gradients generated 
  by the curvature term which cause the network to erroneously change the 
  global brightness, but are quickly recovered from.  
}
\label{fig:loss_mae}
\end{figure}

Figure~\ref{fig:progression} shows how our inpainting algorithm performs
on natural images. After a few thousand iterations, it produces the 
desired result that minimises the MAE. Additional iterations 
still reduce the energy, but deteriorate the MAE and the visual 
impression; see Figure~\ref{fig:loss_mae}. The checkerboard-like 
artefacts in the steady state confirm the previously discussed 
limitations of our discrete energy.
Thus, it is helpful to stop earlier and benefit from the
regularising qualities of the deep image prior. Stopping earlier also
allows us to obtain our results faster. With 
an Nvidia GTX 1080 GPU, it takes $49$ s for the $6,000$ iterations needed 
for the $256 \times 256$ image \textit{trui}.
Finding an automatic stopping criterion is part of our future research.

\medskip
Speed comparisons to other works remain difficult as they often do not 
disclose their runtimes, do not make use of the GPU, or cannot produce 
comparable quality. A rough comparison can be made to the augmented 
Lagrangian approach of Tai et al.~\cite{THC11}. For an $300\times235$ 
image and a fairly dense random mask with $60$ \% known pixels, their 
inpainting takes $78$ s on an Intel Core 2 Duo P8600 @ 2.4 GHz CPU. 
Yashtini and Kang~\cite{YK16} propose an ADMM method and report $854$ s 
for inpainting a $220 \times340$ colour image with a random mask with 
$20$ \% density on a 2.5 GHz Intel Core i5 CPU. These numbers indicate 
that our algorithm offers a competitive speed.

\subsection{Shape Completion}
\begin{figure}
\centering
\footnotesize
\setlength{\tabcolsep}{2.8pt}
\begin{tabular}{ccc}		
\includegraphics[width=27mm, frame]
   {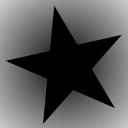} &
\includegraphics[width=27mm, frame]
   {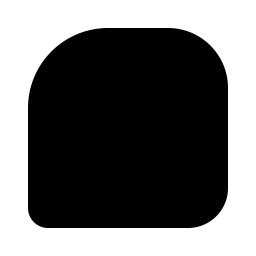} &
\includegraphics[width=27mm, frame]
   {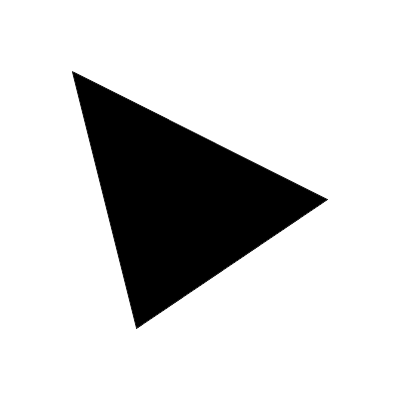} \\[2.8pt]		
\includegraphics[width=27mm, frame]
   {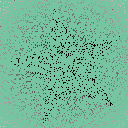} &
\includegraphics[width=27mm, frame]
   {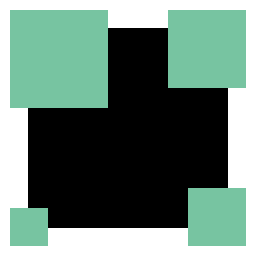} &
\includegraphics[width=27mm, frame]
   {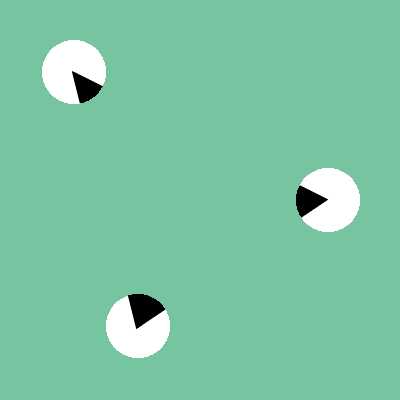}\\[2.8pt]
\includegraphics[width=27mm, frame]
   {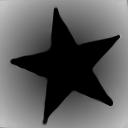} &
\includegraphics[width=27mm, frame]
   {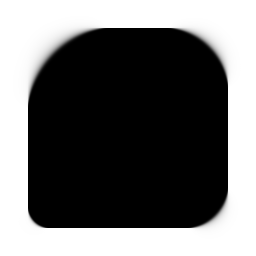} &
\includegraphics[width=27mm, frame]
   {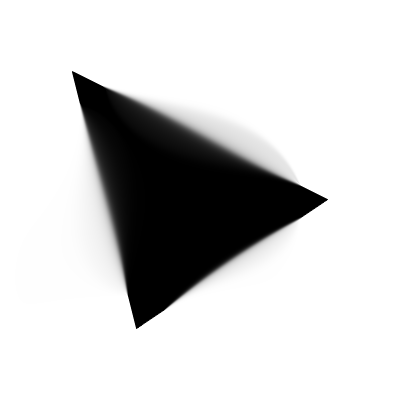} \\[1mm]
$128\!\times\!128$, $b=0.02$, &
$256\!\times\!256$, $b=0.0005$ & 
$400\!\times\!400$, $b=0.0001$,\\
adapted from \cite{THC11} & &
adapted from \cite{SPME14}
\end{tabular}
\caption{Neural inpainting results on shape completion tasks
($\varepsilon=0.0001$). From top to bottom: Original, inpainting domain 
(in green), inpainting result.} 
\label{fig:shape_completions}
\end{figure}
\begin{figure*}
\centering
\footnotesize
\setlength{\tabcolsep}{2.8pt}
\begin{tabular}{cccc}
original & masked image & Chambolle/Pock~\cite{CP19} & 
ours, $b=0.001$\\[1mm]
\includegraphics[width=35mm, frame]
   {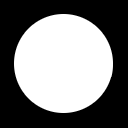} &   
\includegraphics[width=35mm, frame]
   {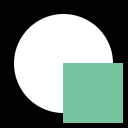} &
\includegraphics[width=35mm, frame]
   {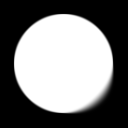} &
\includegraphics[width=35mm, frame]
   {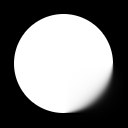} \\[2.8pt]	
\includegraphics[width=35mm, frame]
   {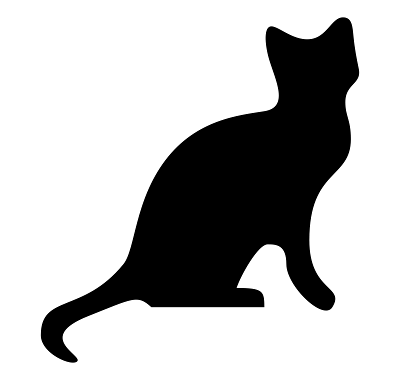} &   
\includegraphics[width=35mm, frame]
   {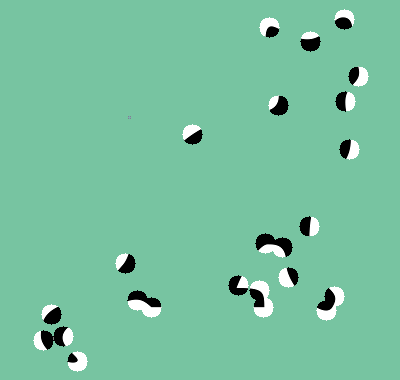} &
\includegraphics[width=35mm, frame]
   {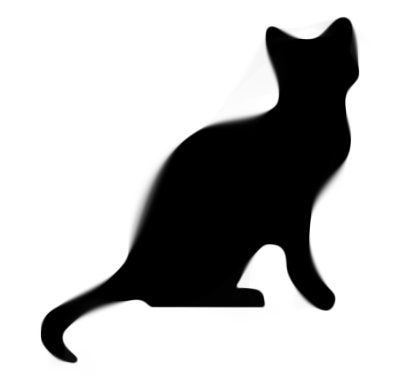} &
\includegraphics[width=35mm, frame]
   {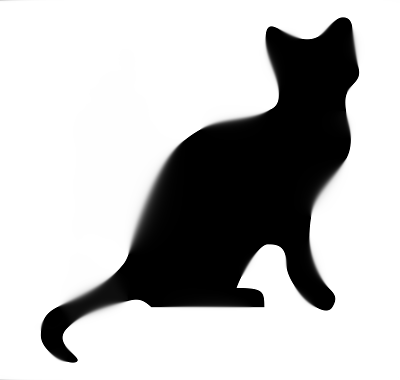} \\[1mm]   
original~\cite{We12} & masked image & Chambolle/Pock~\cite{CP19} &
ours, $b=0.003$
\end{tabular}
\caption{Comparison to the TSC shape completion results of Chambolle and 
Pock~\cite{CP19}, kindly provided by the authors ($\varepsilon=0.0001$). 
Following their experiment, we also initialised the inpainting region of the 
masked image with grey value $0.5$.}
\label{fig:shape_completions_vs}
\end{figure*}
Shape completion allows us to demonstrate that our discretisation produces 
sharp and rotation invariant results. In Figure~\ref{fig:shape_completions}, 
we show that completion of straight edges, curves, and combinations thereof 
are handled adequately. Most noticeably, very large gaps with almost 200 
pixels between the mask regions can be closed. In 
Figure~\ref{fig:shape_completions_vs} we compare to the
state-of-the-art results obtained with the sophisticated lifting approach 
of Chambolle and Pock~\cite{CP19}. While our neural algorithm is simpler,
it produces inpaintings of comparable visual quality.

\section{Conclusions}\label{sec:conclusions}
We have proposed the first neural algorithm for Euler's elastica inpainting.
Research on numerical methods for this attractive but difficult
inpainting model has been pursued for more than two decades and produced 
highly sophisticated techniques. It is thus surprising that our very 
simple approach is qualitatively competitive to a leading elastica 
algorithm for shape completion. 
This speaks for the quality of its parts and the fruitful synergy 
between model-based concepts and deep neural networks. Our finite 
difference approximations for the first and second order derivatives 
of the elastica energy offer good rotation invariance and sharpness. 
Energy minimisation is accomplished automatically by a neural network 
that frees us from the burden of discretising the numerically challenging 
fourth order Euler--Lagrange PDE. Moreover, the regularising properties 
of a deep image prior avoid high-frequent artefacts that the discrete 
energy cannot penalise.  
In contrast to purely data-driven neural approaches, our hybrid
algorithm requires neither ground truth inpaintings nor training
data.

\medskip
Our findings suggests that solvers which combine explicit model 
assumptions with deep energies and deep priors are serious competitors 
in challenging numerical problems. They are particularly useful when 
purely model-based algorithms are difficult to design, too slow, or 
too complex to be analysed. Since these challenges are fairly common 
in advanced scientific computing applications, we expect that the 
impact of such neuroexplicit algorithms will grow.

\bibliography{IEEEabrv,myrefs.bib}

\end{document}